\documentclass[runningheads]{ipi2}
\usepackage{amsmath,amssymb} 
\usepackage{graphicx} 
\usepackage[colorlinks=true]{hyperref}
\hypersetup{urlcolor=blue, citecolor=red}
\usepackage{cases}
\usepackage{bm}
\usepackage{algorithm}
\usepackage{tabularx}
\usepackage{lineno}
\usepackage{color}

\usepackage{xcolor}

\def\currentvolume{abc}
\def\currentissue{def}
\def\currentyear{2020}

\def\ppages{1--2}
\def\DOI{}

\theoremstyle{definition}

\title[Variational Coupling Revisited]
{Variational Coupling Revisited:
	Simpler Models, Theoretical Connections,
	and Novel Applications}

\author[A. Wewior and J. Weickert]{}


\email{wewior@mia.uni-saarland.de}
\email{weickert@mia.uni-saarland.de}

\begin{document}


\maketitle

\centerline{\scshape Aaron Wewior and Joachim Weickert}
\medskip
{\footnotesize
	\centerline{Mathematical Image Analysis Group, }
	\centerline{Faculty of Mathematics and Computer Science, Campus E1.7, 
	Saarland University}
	\centerline{ 66041 Saarbr\"{u}cken, Germany}
} 

%

\bigskip

\begin{abstract}
Variational models with coupling terms are becoming increasingly
popular in image analysis. They involve auxiliary variables, such
that their energy minimisation splits into multiple fractional
steps that can be solved easier and more efficiently. In our paper
we show that coupling models offer a number of interesting properties
that go far beyond their obvious numerical benefits.\\
We demonstrate that discontinuity-preserving denoising can be
achieved even with quadratic data and smoothness terms, provided
that the coupling term involves the $L^1$ norm.
We show that such an $L^1$ coupling term provides additional
information as a powerful edge detector that has remained
unexplored so far.\\
While coupling models in the literature approximate higher
order\linebreak
 regularisation, we argue that already first order coupling
models can be useful. As a specific example, we present a first
order coupling model that outperforms classical TV regularisation.
It also establishes a theoretical connection between
TV regularisation and the Mumford-Shah segmentation approach.
Unlike other Mumford-Shah algorithms, it is a strictly convex
approximation, for which we can guarantee convergence of a
split Bregman algorithm.

\end{abstract}


\section{Introduction}
In image analysis applications, it is desirable to use variational
models with regularisers that involve higher order derivatives:
These higher order models offer better smoothness and avoid
e.g. staircasing artifacts that are present in various first order
variational approaches. Unfortunately, higher order variational
models are numerically unpleasant. For instance, minimising a second
order energy gives rise to a fourth order partial differential
equation (PDE) as Euler-Lagrange equation. Higher order PDEs are more
cumbersome to discretise, and their numerical approximations suffer
from higher condition numbers such that algorithms converge more slowly or
become even unstable.

As a remedy, coupling models have been proposed. They introduce
coupling terms with auxiliary variables. For instance, one
approximates the gradient of the processed image by an auxiliary
variable. In this way a second order energy in a single variable is
replaced by a first order energy in two variables. As Euler-Lagrange
equations, one solves a system of second order PDEs, which is
numerically more pleasant than solving a fourth order PDE.

Coupling models have a long history. Already in 1990, Horn came up
with a coupling model for shape from shading \cite{Ho90}. In the
case of denoising and deblurring, Chambolle and Lions \cite{CL95}
have investigated coupling models
by combining several convex functionals into one regulariser. With
this they were able to modify the total variation regularisation model
\cite{ROF92} and reduce unpleasant staircasing artifacts.
A transfer of this infimal convolution approach to the discrete
setting was proposed by Setzer et al.~\cite{SST11}. Also Bredies 
et al.~\cite{BKP10} performed research in this direction and
introduced a total generalised variation regulariser based on
symmetric tensors of varying order. This allows a selective
regularisation with different smoothness constraints.
More recent coupling models have become highly sophisticated.
Hewer et al.~\cite{HWSSD13} introduced three coupling variables to
come up with a coupling model that replaces fourth order energy
minimisation.
Hafner et al.~\cite{HSW15} considered coupling models that are
anisotropic both in the coupling and the smoothness term, and
Brinkmann et al.~\cite{BBG19} used coupling models in connection with
regularisers based on natural vector field operations, including 
divergence and curl terms.

Nevertheless, in spite of their popularity and sophistication,
the main motivation behind all these coupling approaches was
numerical efficiency. Additional benefits of the coupling
formulation have hardly been investigated so far.

\medskip
{\bf Our Contributions.}
The goal of our paper is to show that the coupling formulation
is much more than just a numerical acceleration strategy. We will
see that it provides valuable benefits from three different
perspectives that have remained unexplored so far:
\begin{itemize}
\item
  From a modelling perspective, we argue that coupling terms may
  free the user from designing nonquadratic smoothness terms if
  discontinuity-preserving solutions are needed. Moreover, we show
  that it can be worthwhile to replace even first order energies by
  coupling models.
\item
  From a theoretical perspective, we demonstrate that coupling models
  may give insights into the connections between two of the most
  widely used variational approaches in image analysis: the
  total variation (TV) regularisation of Rudin et al.~\cite{ROF92}
  and the Mumford-Shah functional \cite{MS89} for image segmentation.
  These insights will also give rise to a new algorithm for
  Mumford-Shah segmentation, that differs from existing ones
  by the fact that it minimises a strictly convex energy. Thus,
  we can guarantee convergence of a split Bregman algorithm.
\item
  From an application perspective, we show that sparsity promoting
  coupling terms may provide useful information as an edge detector
  that comes at no additional cost.
\end{itemize}

\medskip
{\bf Paper Structure.} Our paper is organised as follows:
In Section 2, we derive the framework for a first and second order
model that incorporates the coupling term. We show the relation to the
Mumford-Shah functional and elaborate why the coupling term can be
interpreted as an edge detector. Section 3 deals with the numerical
aspects that lead to the minimisation of our approach in the discrete
setting and showcases the advantages of the split Bregman formulation.
In Section 4, we present experimental results, and we conclude the
paper in Section 5.


\section{Our Models}
In this section we propose two coupling models for edge-preserving 
image
enhancement. We relate them to total variation regularisation and the 
segmentation approach of Mumford and Shah, and we argue that the $L^1$ 
coupling term provides a novel edge detector.
\subsection{Coupling Framework}
\subsubsection{First Order Model.}
Let our image domain $\Omega \subset \mathbb{R}^2$ be an open bounded
set and $f: \Omega \to \mathbb{R}$ denote a greyscale
image that contains noise with zero mean.
Our goal is to find a noise free version of that image. As this is in
general an ill-posed problem, we employ variational approaches to 
find a reconstruction $u$ that is as close as possible to the noise 
free image and at the same time satisfies certain a priori smoothness 
conditions.
One of the earliest models by Whittaker \cite{Wh23} and Tikhonov
\cite{Ti63}
obtains $u$ as minimiser of the energy functional
\begin{equation}
\label{eq:Whittaker-Thikonov}
E(u) = \int_{\Omega} \left(\frac{1}{2} (u-f)^2 + \frac{\alpha}{2}
\|\bm{\nabla} u\|^2 \right) \text{ d}{\bm x} \,,
\end{equation}
where $\|\cdot\|$ denotes the Euclidian norm.
The first term guarantees that $u$ is close to the given data $f$ and
is called \emph{data term}. The second term penalises deviation from 
a smooth solution and is therefore denoted as the \emph{smoothness 
term}.
The regularisation parameter $\alpha > 0$ steers the smoothness of the
solution.
In practice we deal with a rectangular, finite domain $\Omega$. This 
prompts us to choose homogeneous Neumann boundary conditions. These 
coincide with the natural boundary conditions that arise when we derive the 
Euler-Lagrange equations of \eqref{eq:Whittaker-Thikonov}.

We modify \eqref{eq:Whittaker-Thikonov} to a constrained problem by
introducing an auxiliary vector field
 $\bm{v} \in  L^2(\Omega,\mathbb{R}^2)$.
We set $\bm{v} = \bm{\nabla} u$ and obtain
\begin{equation}
\label{eq:Whittaker-Thikonov-constrained}
E(u) = \int_{\Omega} \left(\frac{1}{2} (u-f)^2 + \frac{\alpha}{2}
\|\bm{v}\|^2
\right) \text{ d}{\bm x} \qquad \text{ s.t. }\bm v = \bm \nabla u \,.
\end{equation}
Enforcing the constraint weakly in the Euclidian norm yields the
coupling model
\begin{equation}
\label{eq:Whittaker-Thikonov-coupled}
E(u,\bm v) = \int_{\Omega} \left( \frac{1}{2} (u-f)^2
 + \frac{\alpha}{2}\|\bm{v}\|^2 
 + \frac{\beta}{2} \|\bm{\nabla} u - \bm v\|^2 \right) 
 \text{	d}{\bm x}
\end{equation}
with parameter $\beta > 0$. Since the newly introduced term
inherently couples $\bm \nabla u $ and $\bm v$ we call it
\emph{coupling term}. It is obvious that apart from trivial cases the 
vector field $\bm v$ is not equal to the gradient of 
the function $u$.
However, by following the proof in \cite{BFW17} one can show that
\begin{equation}
\label{eq:equality_in_the_mean}
\int_{\Omega} \bm v \text{ d}{\bm x} 
= \int_{\Omega} \bm \nabla u \text{ d}{\bm x} \,,
\end{equation}
which means that they are "equal in the mean".
The vector field $\bm v$ inherits its boundary conditions from $u$. We will 
discuss the details in the next chapter.

At first sight the coupling model does not offer any advantage over
the energy \eqref{eq:Whittaker-Thikonov} and seems to complicate it
more.
Indeed, by calculating the Euler-Lagrange equations
and performing a simple substitution we see that the minimiser for 
$u$ of the functional in \eqref{eq:Whittaker-Thikonov-coupled} is 
equivalent to the minimiser of energy \eqref{eq:Whittaker-Thikonov} 
with parameter $\frac{\alpha	\beta}{\alpha + \beta}$.
For the limit case $\beta \rightarrow \infty$, which corresponds to a 
hard constraint of $\bm \nabla u = \bm v$, we thus obtain the same 
solution.
In the other limit case $\alpha \rightarrow \infty$, the energy
\eqref{eq:Whittaker-Thikonov-coupled} corresponds to
\eqref{eq:Whittaker-Thikonov} with regularisation parameter $\beta$.

In order to obtain our first order coupling model we only apply a 
small modification to the coupling term in
\eqref{eq:Whittaker-Thikonov-coupled}.
We recall that a function $u \in {L}^1(\Omega)$ is said to be  of
bounded variation ($u \in BV(\Omega)$) if it has a (distributional)
gradient in form of a Radon measure of finite total mass. Then we can
define the total variation seminorm of $u$ as \cite{AFP00}
\begin{equation}
\label{eq:TV_seminorm_cont}
\int_\Omega |\bm \nabla u| :=
\sup \left\lbrace \int_\Omega u \,\text{div} \bm \phi \text{ d}{\bm x}
\,|\, \bm \phi \in \mathcal{C}^1_c (\Omega,\mathbb{R}^2) ,
\, |\bm \phi| < 1\right\rbrace .
\end{equation}
If we consider the difference $\bm \nabla u - \bm v$ as an
argument, we extend \eqref{eq:TV_seminorm_cont} 
and obtain
\begin{equation}
\label{eq:TV_like_seminorm_cont}
\int_\Omega |\bm \nabla u - \bm v| :=
\sup \left\lbrace \int_\Omega u \,\text{div} \bm \phi 
+ \bm v^\top \bm \phi \text{ d}{\bm x}
\,|\, \bm \phi \in \mathcal{C}^1_c (\Omega,\mathbb{R}^2) ,
\, |\bm \phi| < 1\right\rbrace .
\end{equation}
With this we formulate our first order coupling model:
\begin{equation}
\boxed
{
	\label{eq:model_our_first_order}
	E_1(u,{\bm v}) := \frac{1}{2}\int_{\Omega} (u-f)^2 \text{ d}{\bm x}
	+ \frac{\alpha}{2}\int_{\Omega}
	\|{\bm v} \|^2 \text{ d}{\bm x}
	+ \beta \int_{\Omega}
	|{\bm \nabla} u-{\bm v} |\,.
}
\end{equation}
It is obvious that the model \eqref{eq:model_our_first_order} is
strictly convex due to the quadratic data term. Moreover, we also
have a quadratic smoothness term and only the coupling term
contains a more sophisticated penaliser. It is therefore no surprise 
that for the hard constraint $\beta \rightarrow \infty$,
the energies \eqref{eq:model_our_first_order} and
\eqref{eq:Whittaker-Thikonov-coupled}
are still identical. The more interesting case arises in the
limit $\alpha \rightarrow \infty$: In order to fulfil the constraint, 
the
vector field $\bm v$ should be zero, which yields the regulariser
\eqref{eq:TV_seminorm_cont} for the coupling term. Energy
\eqref{eq:model_our_first_order} is then equivalent to the classical
TV denoising model by Rudin et al.~\cite{ROF92}. However, we will
later demonstrate how the structure of the coupling model is both
beneficial for an efficient minimisation as well as for superior
denoising results.
We note that it can be shown that model 
\eqref{eq:model_our_first_order} is equivalent to denoising with a Huber TV 
penaliser \cite{BP16}.

\subsubsection{Second Order Model.}
While TV regularisation promises a good compromise between image
reconstruction and  edge preservation, it introduces "blocky"
structures into the reconstructed images, the so-called staircasing
artifacts.
One way to avoid these artifacts is by second order models
\cite{Sche98,YK00,LLT03}. Due to the coupling formulation, a higher
regularisation order can easily be realised in our framework.
Instead of the vector field $\bm v$ we penalise its Jacobi matrix in
the Frobenius norm. The resulting energy is given by
\begin{equation}
\boxed
{
	\label{eq:model_our_second_order}
	E_2(u,{\bm v}) := \frac{1}{2}\int_{\Omega} (u-f)^2 \text{ d}{\bm 
	x}
	+ \frac{\alpha}{2}\int_{\Omega}
	\|{\bm \nabla \bm v} \|_F^2 \text{ d}{\bm x}
	+ \beta \int_{\Omega}
	|{\bm \nabla} u-{\bm v} |\,.
} 
\end{equation}
In order to see that \eqref{eq:model_our_second_order} approximates a
second order model, we discuss the limiting case
$\beta \rightarrow \infty$ again.
The smoothness term favours a constant vector field $\bm v$. For a
hard coupling this means that constant first derivatives of $u$ lead
to the smallest energy. Hence, the solution $u$ corresponds to an
affine function which is characteristic for second order
regularisation methods.
The model resembles the \emph{total generalized variation (TGV)} of
second order \cite{BKP10}:
\begin{equation}
\label{eq:TGV}
E_{TGV}(u,{\bm v}) := 
\frac{1}{2 }\int_{\Omega} (u-f)^2 \text{ d}{\bm x}
+ \alpha\int_{\Omega} |{\bm{\varepsilon}(\bm v)} |
+ \beta \int_{\Omega} |{\bm \nabla} u-{\bm v} |
\end{equation}
where $\bm{\varepsilon}$ denotes the symmetrised gradient in a TV-like
norm in the smoothness term.
However, the smoothness term in our first and second order model is
still quadratic.
In Section 3 we will see that this allows very efficient algorithms.

\subsection{Relation to the Segmentation Model of Mumford and Shah}
\label{sec:review_MS}
Let $\Omega$ be defined as before and let $K$ be a compact curve in
$\Omega$. The segmentation model by Mumford and Shah minimises
the following energy \cite{MS89}:
\begin{equation}\label{eq:mumford_shah}
E_{MS} (u, K) =
\frac{1}{2}\int_{\Omega}(u-f)^2 \text{ d}{\bm x}
+ \frac{\alpha}{2} \int_{\Omega \backslash K} \|\bm \nabla u\|^2
\text{ d}{\bm x}
+ \beta\, \ell(K) \,.
\end{equation}
The function $u :\Omega \rightarrow \mathbb{R}$ approximates
$f$ as a sufficiently smooth function in $\Omega \backslash K$,
but admits discontinuities across $K$. The curve $K$ denotes
the edge set between the segments. The expression $\ell(K)$ represents
the length of $K$ and can be expressed as $\mathcal{H}^1(K)$,
the one-dimensional Hausdorff measure in $\mathbb{R}^2$.
The parameter $\beta >0 $ controls the length of the edge set,
whereas  $\alpha >0 $ steers the smoothness of $u$ in
$\Omega\backslash K$.

Unfortunately the Mumford-Shah energy is highly non-convex and difficult to 
minimise.
Ambrosio and Tortorelli \cite{AT92} showed that they
can approximate the solution by a sequence of regular functions defined on
Sobolev spaces and proved $\Gamma$-convergence.
However, their energy is still non-convex and the quality of the solution
depends on the initial parameter setting of the $\epsilon$-approximation.

Cai et al.~\cite{CCZ13} proposed a variant that approximates the
edge length by the total variation seminorm. They were able to
approximate \eqref{eq:mumford_shah} by a convex minimisation problem
over the whole domain:
\begin{equation}\label{eq:model_cai}
E_{\text{Cai}}(u) = \frac{1}{2}\int_{\Omega}(u-f)^2 \text{ d}{\bm x}
+ \frac{\alpha}{2} \int_{\Omega } \|\bm \nabla u\|^2 \text{ d}{\bm x}
+ \beta \int_{\Omega} |\bm \nabla u|
\,.
\end{equation}
Nevertheless, the price for convexity comes with a smooth minimiser
$u \in W^{1,2}(\Omega)$ that cannot contain any discontinuities such
that a postprocessing step is ne\-cessary for a segmentation.

In order to relate equation \eqref{eq:model_cai} to our first order 
model 
\eqref{eq:model_our_first_order}, we perform the image decomposition $u = 
u_1 + u_2$ such that the regularisers affect different parts of the image. 
These kind of approaches based on an infimal convolution were introduced by
Chambolle and Lions in \cite{CL95}. With this we obtain
\begin{equation}\label{eq:model_decomposed}
E(u_1,u_2) = \frac{1}{2}\int_{\Omega}(u_1+u_2-f)^2 \text{ d}{\bm x}
+ \frac{\alpha}{2} \int_{\Omega }
\|\bm \nabla u_1\|^2 \text{ d}{\bm x}
+ \beta \int_{\Omega} |\bm \nabla u_2|
\,.
\end{equation}
The decomposition lets us define the parts on different spaces.
While $u_1 \in W^{1,2}(\Omega)$, we have $u_2 \in BV(\Omega)$.
Thus, we admit jumps in $u$ which makes a postprocessing unnecessary.
By introducing an auxiliary vector field $\bm v \in 
L^2(\Omega,\mathbb{R}^2)$ which is designed to
approximate $\bm \nabla u_1$, we consequently obtain for the
argument of the last term:
\begin{equation}
\bm \nabla u_2 = \bm \nabla u - \bm \nabla u_1
\approx \bm \nabla u - \bm v
\,.
\end{equation}
Substituting this in \eqref{eq:model_decomposed} brings us back 
to the coupling model \eqref{eq:model_our_first_order}.
Therefore, we observe that our first order coupling model 
approximates the Mumford-Shah segmentation model.
It is obvious that this convex approximation is smoother then other 
variants as we do not perform an enhancement at edges.
However, we still allow jumps in the regularised image.
Moreover, we stress that the approximation is strictly convex and admits a 
unique solution.

The reasoning for the first order model can easily be transferred to 
the second order model. Energy \eqref{eq:model_our_second_order} can 
thus be understood as an approximation to a second order 
Mumford-Shah model.
One example for a second order approximation is the model of Blake and 
Zisserman \cite{BZ87}:
\begin{linenomath}
	\begin{align}
	\label{eq:mumford_shah_Blake_Zisserman}
	\begin{split}
	E_{BZ} (u, K_0, K_1) =
	\frac{1}{2}\int_{\Omega}(u-f)^2 \text{ d}{\bm x}
	+ \frac{\alpha}{2} \int_{\Omega \backslash (K_0 \cup K_1)} \|\bm \nabla^2 
	u\|^2	\text{ d}{\bm x}\\
	+ \beta\, \mathcal{H}^1(K_0) + \gamma\, \mathcal{H}^1(K_1) \,.
	\end{split}
	\end{align}
\end{linenomath}
Here, $\bm \nabla^2$ denotes the Hessian, and the Hausdorff measures on the 
curves $K_0$ and $K_1$ penalise not only the edge length between segments 
as before, but also the length of the curve that describes creases in the 
image.
More recent variants were discussed in \cite{BE15,DDPYB15}. They all have 
in common that they require the approach of Ambrosio-Torterelli to derive a 
minimiser. Our second order model gives an alternative that manages without 
the $\epsilon$-approximation.
\subsection{Interpretation of the Coupling Term as an Edge Detector}
In the previous subsection we established a connection between our
coupling framework and the Mumford-Shah functional. The difficulty in
the original formulation lies in the computation of the edge set $K$.
We argue in the following how the edge set is related to the
introduced coupling term which is directly accessible without
additional effort.

We observe that the smoothness terms in energy
\eqref{eq:model_our_first_order} and \eqref{eq:model_our_second_order}
are quadratic. That means that on one hand
$\bm v $ is still a smooth approximation to the $\bm \nabla u$.
On the other hand the coupling term contains the sparsity-promoting
$L^1$ norm. As a result, $\bm v$ deviates from $\bm \nabla u$ only at
a small number of positions, which are represented as peaks in the
energy of the coupling term. Since these peaks correspond to
discontinuous jumps in the solution $u$, we see that the coupling
term corresponds to the desired edge set and can naturally be
utilised as an edge detector.

We distinguish two limiting cases.
For $\beta \rightarrow \infty$ we end up with the hard constraint
$\bm \nabla u = \bm v$ such that the coupling energy
\eqref{eq:TV_like_seminorm_cont} evaluates to $0$.
This makes sense, as the model simplifies to energy
\eqref{eq:Whittaker-Thikonov} again, and $u$ will be a smooth solution
without jumps.
For $\beta \rightarrow 0$, the functions $u$ and $\bm v$ are
decoupled.
A minimiser is reached when $u$ corresponds to the given image $f$
and $\bm v =\bm 0$.
Here the coupling term represents a simple gradient-based edge
detector without a presmoothing. Such a detector is not reliable in 
the presence of high frequency oscillations and noise as it responds
to many false edges.

In between these limiting cases, an increasing value for $\beta$ 
allows the coupling term to partition the image into more and more 
homogeneous regions, since this gives a smaller energy.
Thus, on each scale $\beta$, we solve a convex variational model and
obtain a unique solution. The coupling term serves as a
novel edge detector that provides us with a globally optimal edge set.
The discontinuity-preserving $L^1$ norm in the underlying model 
allows the coupling term to respond at edges, but makes it robust 
against false edges produced by noise in the image.
As it also promotes sparsity it limits the detector's response
to the most relevant edges which are well localised.


\section{Numerical Aspects}
In this section we argue that the discrete variables in our models 
live on different grids in order to obtain consistent approximations.
We speed up a conventional alternating minimisation by a split Bregman 
scheme \cite{GO09} that offers a tremendous acceleration of the 
minimisation process. We discuss the scalar-valued problem and then 
explain 
the modification needed to solve the problem for vector-valued data.
\subsection{Discretisation of the Variables}
In order to implement the model on a computer we introduce a discrete 
setting
based on a finite difference discretisation. We consider the domain 
$\Omega$ to
be rectangular, i.e. $\Omega = (a,b) \times (c,d)$. Then the discrete 
domain 
$\Omega_d$ is given on a regular Cartesian grid with $M \times N$ nodes:
\begin{equation}
\Omega_d := \left\lbrace 
\left( a+ \left( i-\tfrac{1}{2}\right) h,
c+ \left( j-\tfrac{1}{2}\right) h \right):
1\leq i \leq M,\, 1 \leq j \leq N\right\rbrace
\end{equation}
with grid size $h := \frac{b-a}{M} = \frac{d-c}{N}$.
The discrete locations $(i,j)$ within 
$\Omega_d$ correspond w.l.o.g. to the continuous locations $\left( a+ 
\left( i-\frac{1}{2}\right)h,\, c+ \left( j-\frac{1}{2}\right) h \right)$ 
within $\Omega$.
Discretising the scalar-valued functions $u$ and $f$ at positions
$(i,j)$ yields variables $\bm{u}, \bm{f} \in \mathbb{R}^{M \times N}$.
We approximate the derivatives with a finite difference scheme.
As we require the consistency order of our approximation to be at 
least two, we use central differences around a grid shifted by 
$\frac{h}{2}$.
In order to be consistent with this approximation we consequently need
the discretisation for $\bm v = ({\bm v_1}, {\bm v_2})^\top$ to live 
on different dual grids shifted just by this half grid size \cite{HS97}.
More precisely, entries $(v_1)_{i,j}$ are defined at positions
$(i+\frac{1}{2},j)$, whereas entries $(v_2)_{i,j}$ are defined
at positions $(i,j+\frac{1}{2})$, as they approximate the
partial derivatives in different directions. 
This also extends to the boundary conditions.
As already mentioned $\bm v$ inherits its boundary conditions from the 
homogeneous Neumann boundary conditions of $\bm u$.
This means that the components of $\bm v$ have zero Dirichlet boundary 
conditions in the direction of the shifted grid and homogeneous Neumann 
boundary conditions otherwise.

\subsection{Discretisation of the Energies}
For sufficiently smooth data $u$, we can define a regularised 
penaliser of \eqref{eq:TV_seminorm_cont} by 
\begin{linenomath}
\begin{equation}
\label{eq:penaliser_cont}
\int\limits_\Omega \varphi_\epsilon( \bm \nabla u) \text{ d}{\bm x} 
= \int\limits_\Omega \sqrt{\|\bm \nabla u \|^2 + \epsilon} 
\text{ d}{\bm x}
\end{equation}
\end{linenomath}
with small $\epsilon  \geq 0$ \cite{AV94}.
For the case $\epsilon =0$, it reduces to \eqref{eq:TV_seminorm_cont},
but otherwise offers the advantage to be differentiable when $\bm 
\nabla u = 0$. We consider this regularisation for our discrete 
implementations.

The discretisation of our first order model
\eqref{eq:model_our_first_order} and second order model
\eqref{eq:model_our_second_order} subsume to the minimisation
problem
\begin{linenomath}
\begin{equation}
\label{eq:model_our_discrete}
\boxed{
	\min\limits_{\bm u, \bm v} \left\lbrace \frac{1}{2} \| \bm u - \bm
	f\|^2_2
	 +	\frac{\alpha}{2} \,S^\ell(\bm v)
 + \beta \, C(\bm \nabla_{\!\!u} \bm u - \bm	 v) \right\rbrace
}
\end{equation}
\end{linenomath}
where $\| \cdot \|_2$ denotes the discrete Euclidean norm.
The coupling term 
contains the discrete counterpart of the regularised
penalising function \eqref{eq:penaliser_cont} in form of
\begin{linenomath}
\begin{equation}
C\left(\bm \nabla_{\!\!u} \bm u - \bm v\right) = \sum_{i=1}^{M} \sum_{j=1}^{N}
\varphi_\epsilon(\|\bm \nabla_{\!\!u} \bm u - \bm	 v\|_2^2)_{i,j}
\end{equation}
\end{linenomath}
which evaluates the regulariser \eqref{eq:penaliser_cont} pixelwise with
\begin{linenomath}
\begin{align}\label{eq:TV_seminorm}
\begin{split}
\varphi_\epsilon( \|\bm \nabla_{\!\!u} \bm u {-} \bm
v\|_2^2)_{i,j} =
\Bigg(\frac{1}{2}
&\bigg(
\left.(\partial_{x+} \bm u - v_1)^2\right|_{i,j}
+ \left.(\partial_{x+} \bm u - v_1)^2\right|_{i-1,j}\\[-2.5ex]
+&\left.(\partial_{y+} \bm u - v_2)^2\right|_{i,j}
+ \left.(\partial_{y+} \bm u - v_2)^2\right|_{i,j-1}\bigg)
+ \epsilon
\Bigg)^{\frac{1}{2}}\!.
\end{split}
\end{align}
\end{linenomath}
The approximation of the discrete gradient components 
$\partial_{x+} \bm u$ and $\partial_{y+} \bm u$ lives on a grid 
shifted by half a grid size in the corresponding positive 
coordinate direction and incorporates 
the homogeneous Neumann boundary conditions.

In the function $S^\ell(\bm v)$, the parameter $\ell$ defines the 
regularisation order of the problem.
For $\ell=1$, we have $S^1(\bm v) = \| \bm v \|^2_2$, and for $\ell=2$
we use $S^2(\bm v)$ given by
\begin{linenomath}
\begin{equation}
\| \bm \nabla_{\!\!v}
\bm v \|^2_2 := \sum_{i=1}^{M} \sum_{j=1}^{N} \left( (\partial_{x-} \bm
v_1)^2_{i,j} + (\partial_{y+} \bm v_1)^2_{i,j} + (\partial_{x+} \bm
v_2)^2_{i,j} + (\partial_{y-} \bm v_2)^2_{i,j} \right)\,.
\end{equation}
\end{linenomath}
We employ the same discretisation for $\partial_{y+} \bm v_1$ and 
$\partial_{x+} \bm v_2$ as above,
whereas we introduce the approximations $\partial_{x-} \bm v_1$
and $\partial_{y-} \bm v_2$ as central differences on a grid
which is shifted in the respective negative coordinate
 with Dirichlet boundary conditions.

If we solve problem \eqref{eq:model_our_discrete} in a naive way,
we set the derivatives with respect to the components $\bm u$ and
$\bm v$ to zero and perform an alternating iterative minimisation.
The nonlinearity is always evaluated at the old
iterate. The minimisation problem for $\bm u$ leads to
\begin{linenomath}
\begin{equation}
\label{eq:ELE_u}
\bm 0 =\bm u- \bm f
- \beta \textrm{ div}_{\!u}\left( \varphi_\epsilon'(\|\bm 
\nabla_{\!\!u}
 \bm u^k - \bm	v^{k}\|_2^2)
 (\bm \nabla_{\!\!u} \bm u - \bm v^k) \right)
\end{equation}
\end{linenomath}
with the diffusivity vector
$\varphi_\epsilon'(\|\bm \nabla_{\!\!u}  \bm u^k - \bm	v^k\|_2^2) =
\frac{1}{\sqrt{\|\bm \nabla_{\!\!u}  
\bm u^{k} - \bm	v^k\|^2+\epsilon}}$ and a discrete divergence 
operator $\textrm{div}_{\!u}$.
For $\ell = 1 $, the auxiliary vector field $\bm v$ requires us to 
solve the following equations:
\begin{linenomath}
\begin{align}
\label{eq:ELE_v_1}
\bm 0 & = \varphi_\epsilon'(\|\bm \nabla_{\!\!u}  \bm u^{k} - \bm	
v^k\|_2^2)
(\bm v - \bm \nabla_{\!\!u} \bm u^{k+1}) + \alpha \bm v ,
\end{align}
\end{linenomath}
whereas for $\ell =2$ we obtain a system of equations
of the form
\begin{linenomath}
\begin{align}
\label{eq:ELE_v_2}
\bm 0 & = \varphi_\epsilon'(\|\bm \nabla_{\!\!u}  \bm u^k - \bm	
v^k\|_2^2)
( \bm	v - \bm \nabla_{\!\!u}  \bm u^{k+1})
- \alpha \bm \Delta_v \bm v .
\end{align}
\end{linenomath}
Here, $\bm \Delta_v$ denotes a direct sum of discrete Laplace 
operators that are applied to each component of $\bm v$ separately.

\subsection{The Split Bregman Method}
Unfortunately, the solution of \eqref{eq:ELE_u} is a bottleneck
in the alternating minimisation process. The derivatives of the
regularised penalisation that serve as a diffusivity slow down the
convergence speed drastically when we apply iterative solvers.
As a remedy for this problem we employ the split Bregman
method by Goldstein and Osher \cite{GO09}.
This operator splitting approach is equivalent to the 
alternating direction method of multipliers (ADMM) and the 
Douglas-Rachford splitting algorithm which are suitable for convex 
optimisation problems such as \eqref{eq:model_our_discrete} and whose 
convergence is guaranteed \cite{BN10,Se09}.

We introduce the discrete vector field $\bm p$ that lives on the same
grid as $\bm v$. We set $\bm p := \bm \nabla_{\!\!u} \bm u - \bm v$
 and obtain the constrained problem
 \begin{linenomath}
\begin{equation}
\min\limits_{\bm u, \bm v, \bm p}\left\lbrace\frac{1}{2} \|\bm u - \bm
f\|^2_2 
+ \frac{\alpha}{2} \,S^\ell(\bm v)
+ \beta \,C\left(\bm p\right) 
\right\rbrace \quad
\text{s.t. }
\bm p = \bm \nabla_{\!\!u} \bm u - \bm v  .
\end{equation}
\end{linenomath}
Enforcing the constraint strictly with a numerical parameter
 $\lambda > 0$ yields the split Bregman iteration at step $k+1$:
 \begin{linenomath}
\begin{align}
\label{eq:minimisation_problem_uvp}
\begin{split}
(\bm u^{k+1}, \bm v^{k+1}, \bm p^{k+1})  = {} &
 \arg\min\limits_{\bm u, \bm v,	\bm	p}
\left\lbrace
\frac{1}{2} \|\bm u - \bm f\|^2_2 
+ \frac{\alpha}{2} \,S^\ell(\bm v) 
+ \beta \, C\left(\bm p\right) \right.  \\
& \hspace*{4.5em}\left. + \, \frac{\lambda}{2} \| \bm
p- \bm \nabla_{\!\!u} \bm u + \bm v - \bm b^k \|^2_2
\right\rbrace,
\end{split}\\
\label{eq:update_b}
\bm b^{k+1} =
 {} & \bm b^k + \bm \nabla_{\!\!u} \bm u^{k+1} - \bm v^{k+1} -
\bm p^{k+1}.
\end{align}
\end{linenomath}
Even though the split Bregman method introduces an additional
minimisation problem, the previous nonlinear equations are simplified
substantially due to the simple quadratic structure of the data
and smoothness term.
The subproblems for $\bm u$ and $\bm v$ are then by design
linear and yield the optimality conditions
\begin{linenomath}
\begin{align}
\label{eq:optimality_condition_u}
\left(\bm I - \lambda \bm \Delta_u\right) \bm u \,\,=\,\, 
& \bm f + \lambda \textrm{ div}_{\!u}
\left(\bm p^k + \bm v^k - \bm b^k\right)\,,\\
\label{eq:optimality_condition_v_1}
\bm v  \,\,=\,\,  &\frac{\lambda}{\alpha + \lambda} 
\left(\bm \nabla_{\!\!u} \bm u^{k+1} - \bm p^k + \bm b^k \right) 
& \text{for }\ell =1, \\
\label{eq:optimality_condition_v_2}
\left(\lambda \bm I - \alpha \bm \Delta_v \right) \bm v  \,\,=\,\,  
& \lambda \left(\bm \nabla_{\!\!u} \bm u^{k+1} - \bm p^k + \bm b^k
\right)
&\text{for }\ell =2.
\end{align}
\end{linenomath}
Equation \eqref{eq:optimality_condition_v_1} leads to a 
simple pointwise computation.
For the other two subproblems we may only apply a few steps of an 
iterative solver. This is in accordance to Osher and Goldstein who state 
that a complete convergence of the subproblems is not necessary for the 
convergence of the whole algorithm \cite{GO09}.

The last subproblem for $\bm p$ contains the nonlinearity, that now
only occurs pointwise and not as a diffusivity in a divergence term.
We obtain the update rule
\begin{linenomath}
\begin{align}\label{eq:update_p}
\bm p^{k+1}  =
\frac{\lambda}{\lambda + \varphi_\epsilon'(\| \bm p^k\|_2^2)}
\left( \bm \nabla_{\!\!u} \bm u^{k+1} - \bm v^{k+1} + \bm 
b^k\right).
\end{align}
\end{linenomath}
We terminate the algorithm if the residuum of equations 
\eqref{eq:ELE_u}--\eqref{eq:ELE_v_2} in relation to an initial 
residuum gets smaller than a prescribed threshold.
If we are also interested in the resulting edge set, we can obtain it
without additional effort by evaluating $\|\bm \nabla_{\!\!u} \bm u - 
\bm v\|_2^2$.

\subsection{Vector-valued Extension}
An extension to vector-valued images is straightforward. The
Euclidian norm in data and smoothness terms allows a separate
calculation for each channel which opens up the possibility for a
parallel implementation.
The only coupling between channels occurs in \eqref{eq:TV_seminorm}.
Therefore, only the nonlinearities in the pointwise update  rule
\eqref{eq:update_p} are affected.


\section{Experiments}

\subsection{Denoising Experiment}
\begin{figure}[!ht]
	\centering
	\begin{tabular}{ccc}
		\includegraphics[width=0.42\textwidth]
		{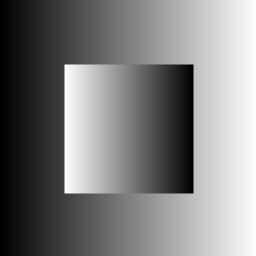} &		
		\hspace{0.5cm}$\,$&
		\includegraphics[width=0.42\textwidth]
		{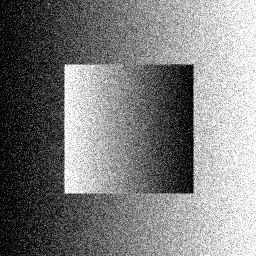}\\
		test image  \cite{BBG19} 
		&& 
		\begin{tabular}{@{}c@{}}
			with Gaussian noise \\ ($\sigma =40$)
		\end{tabular}\\[3ex]
		\includegraphics[width=0.42\textwidth]
		{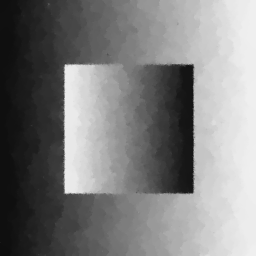} &		
		\hspace{3cm}&
		\includegraphics[width=0.42\textwidth]
		{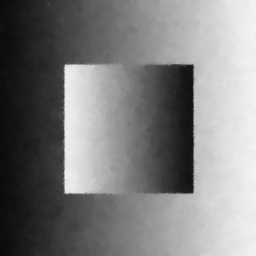} \\
		TV (MSE: 35.06) & 
		&
		\begin{tabular}{@{}c@{}}
			ours, 1st order \\ (MSE: 31.79)
		\end{tabular} \\[3ex]
		\includegraphics[width=0.42\textwidth]
		{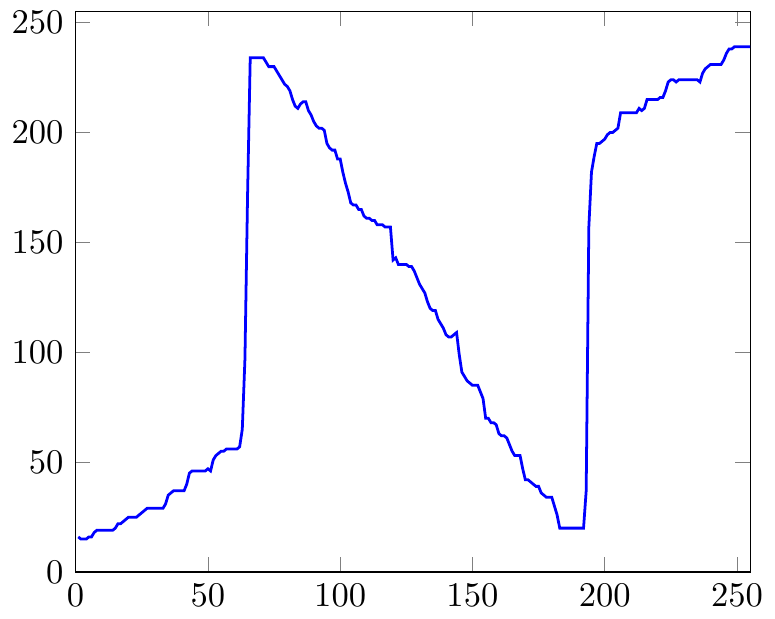}&	
		\hspace{3cm}&
		\includegraphics[width=0.42\textwidth]
		{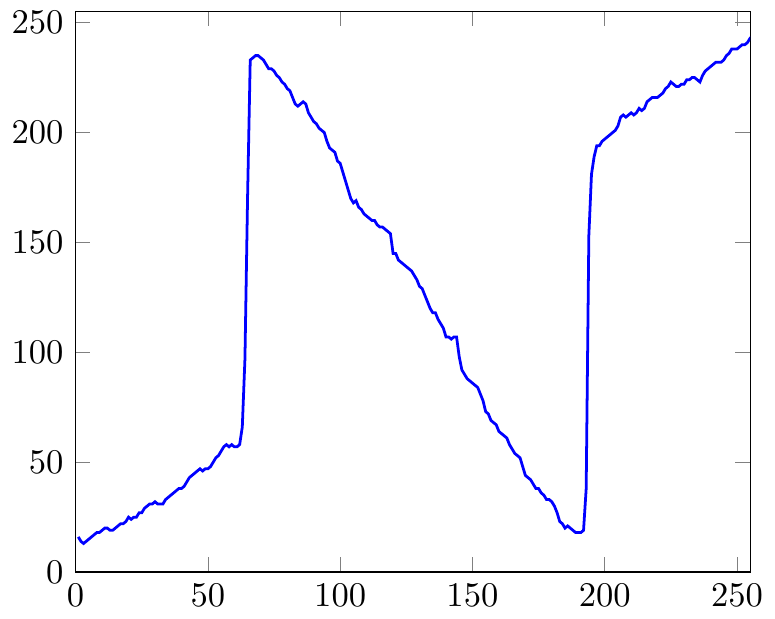}\\
		row 87, TV model & & 		
		\begin{tabular}{@{}c@{}}
			row 87, 1st order model
		\end{tabular}
	\end{tabular}
	\caption{Comparison of denoising result for first order models}
	\label{fig:figure1}
\end{figure}
In this section we discuss the application of our first and second
order coupling models for denoising.
We compare our models to TV denoising \cite{ROF92} and the combined 
first and second order approach TV-TV$^2$ by Papafitsoros et 
al.~\cite{PS14}.
 We set the regularisation parameter of the penalising function to 
 $\epsilon = 10^{-6}$ and stop the algorithms if the relative 
 residuum  gets smaller than $10^{-6}$. For the solution of the 
 linear subproblems we perform an inner loop of $10$ Jacobi steps per
 iteration.
  The numerical parameter $\lambda$ is important for the number of
 iterations until convergence is reached. Empirically we found that
 $\lambda$ should be chosen in dependence of  $\beta$ and the 
 regularisation order $\ell$. We fix $\lambda := \ell \cdot \beta $.
We choose the parameters optimally for each method such that the mean
squared error (MSE) w.r.t. the ground truth is minimised.

In Fig. \ref{fig:figure1} we consider an affine
test image of size $256\times 256$ that is corrupted by Gaussian 
noise with $\sigma = 40$.
We first consider denoising by first order models.
	Our model is able to beat the classical TV model w.r.t. the MSE.
	We also see a qualitative improvement when we consider a horizontal cross 
	section of the images in the last row. We reduce the staircasing 
	artifacts 
	significantly.

\begin{figure}[hbt]
	\centering
	\begin{tabular}{ccc}
		\includegraphics[width=0.42\textwidth]
		{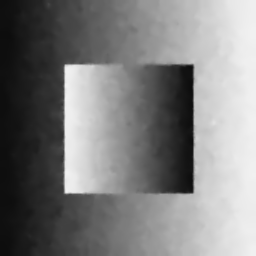}
		&\hspace{0.5cm}$\,$&
		\includegraphics[width=0.42\textwidth]
		{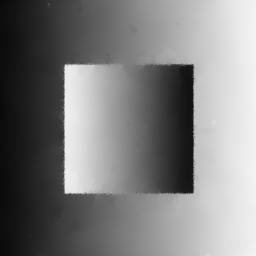} \\
		\begin{tabular}{@{}c@{}}
			TV-TV$^2$ \\
			(MSE: 31.35)
		\end{tabular}	
		& &
		\begin{tabular}{@{}c@{}}
			ours, 2nd order\\ (MSE: 26.02)
		\end{tabular} \\[3ex]
		\includegraphics[width=0.42\textwidth]
		{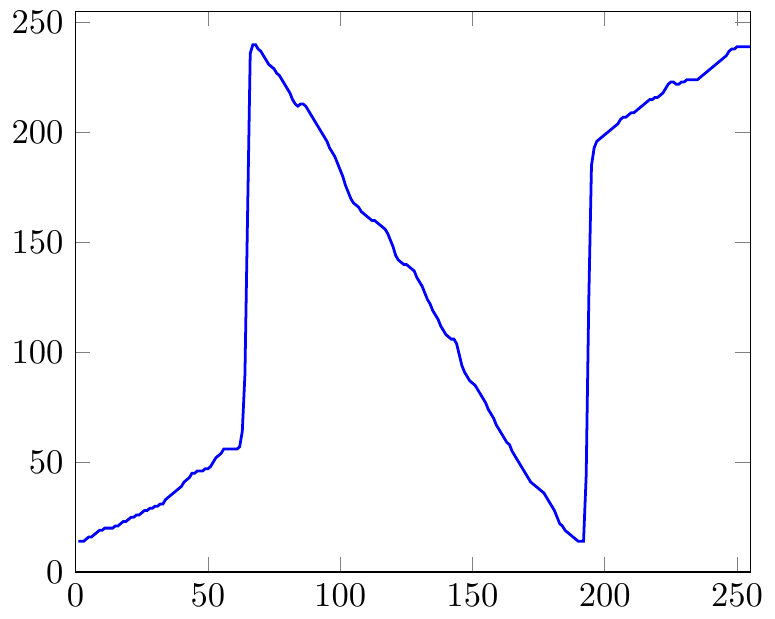}&	
		\hspace{3cm}&
		\includegraphics[width=0.42\textwidth]
		{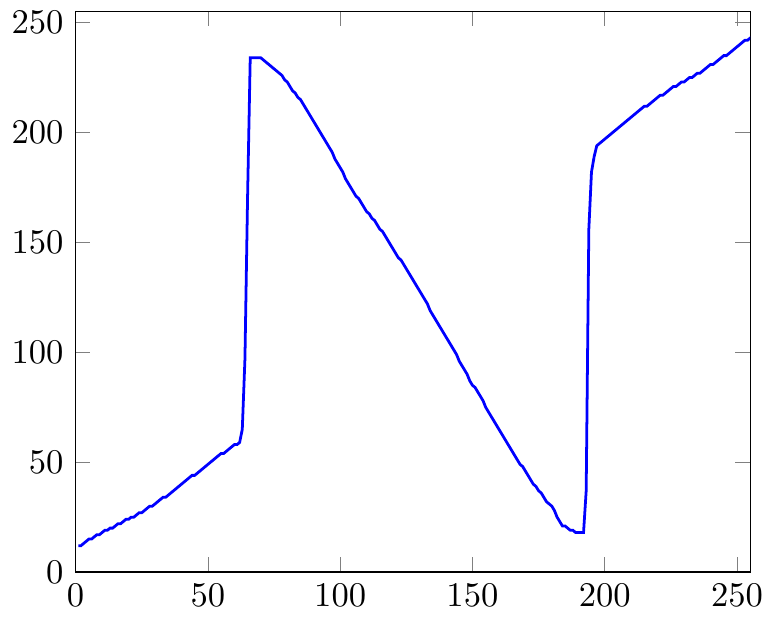}\\
		row 87, TV-TV$^2$ model & & 		
		\begin{tabular}{@{}c@{}}
			row 87, 2nd order model
		\end{tabular}
	\end{tabular}
	\caption{Comparison of denoising result for second order models}
	\label{fig:figure2}
\end{figure}

We compare the reconstruction by second order models in Fig. 
\ref{fig:figure2}.
While the TV-TV$^2$ model beats our first order coupling model only 
slightly w.r.t. to the MSE, our second order model presents a large 
improvement. In the 1D plot we see that no staircasing artifacts
arise.

This experiment also illustrates the efficiency of the split Bregman  
implementation for solving our problems.
We compare two implementations. 
	On the one hand we employ an alternating minimisation strategy of 
	equations \eqref{eq:ELE_u} and \eqref{eq:ELE_v_1} resp. 
	\eqref{eq:ELE_v_2} realised with a standard conjugate gradient method.
	On the other hand we apply a split Bregman Implementation based on the 
	subproblems that arise from equation \eqref{eq:minimisation_problem_uvp}.

	\begin{figure}[htb]
		\centering
		\begin{tabular}{cc}
			\includegraphics[width=0.45\textwidth]
			{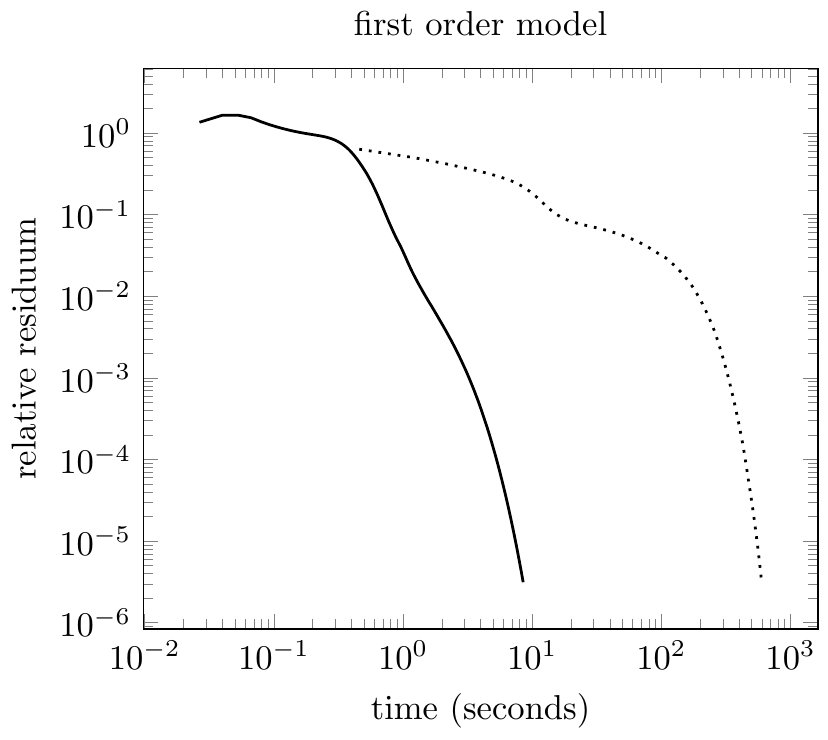} &	
			\includegraphics[width=0.45\textwidth]
			{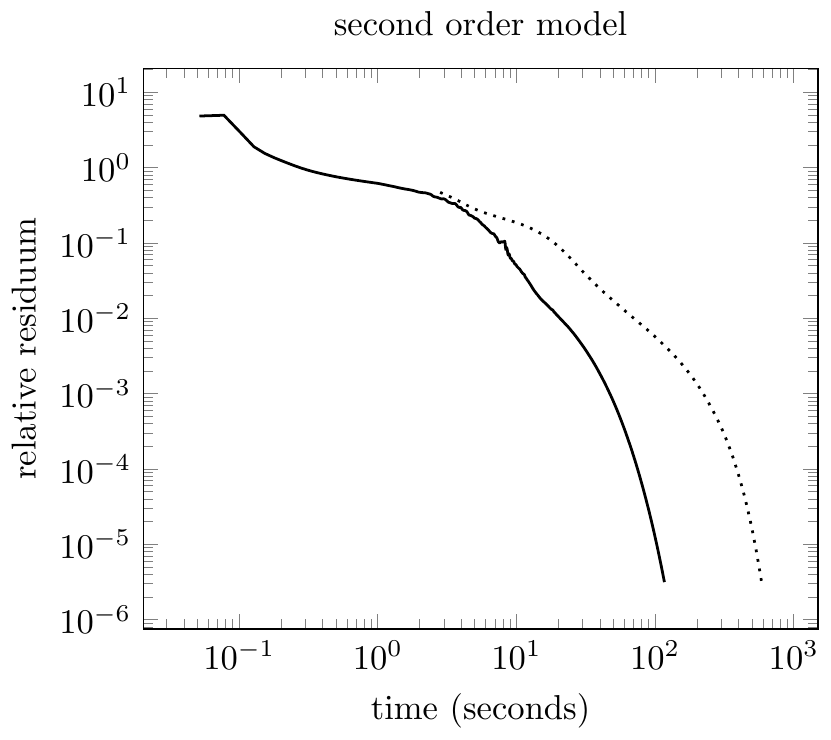} 
		\end{tabular}
		\caption{A log-log plot of runtime comparison of implementation
			with conjugate 	gradient scheme (dotted) and split Bregman
			implementation (straight line).}
		\label{fig:figure3}
	\end{figure}
	
Figure \ref{fig:figure3} shows the decay in the relative residuum 
as a function to the runtime. 
For both the first and second order model, we gain a speed up of more 
than one order of magnitude with the split Bregman scheme.

\subsection{Convex Mumford-Shah Approximation}
To understand how the coupling model approximates the Mumford-Shah
functional we first have to discuss the influence of the parameters.
The parameter $\beta$ determines the smoothness \emph{between}
segments.
For a fixed value of $\alpha$, increasing $\beta$ reduces the number
of segments and consequently of detected edges. To counteract the
segment fusion steered by $\beta$, the parameter $\alpha$ should be chosen
relatively large to benefit from the TV-like effect \emph{inside}
each segment.

\begin{figure}[htb]
	\centering
	\begin{tabular}{ccc}
		\includegraphics[width=0.3\textwidth]
		{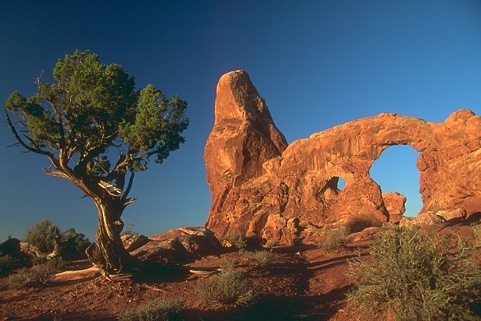} &
		\includegraphics[width=0.3\textwidth]
		{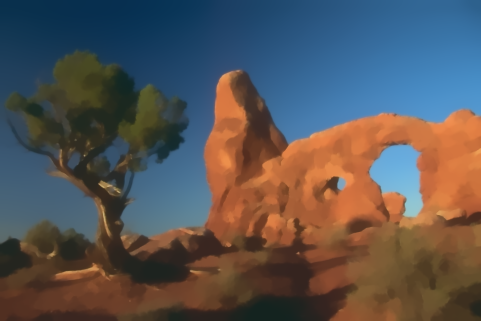} &
		\includegraphics[width=0.3\textwidth]
		{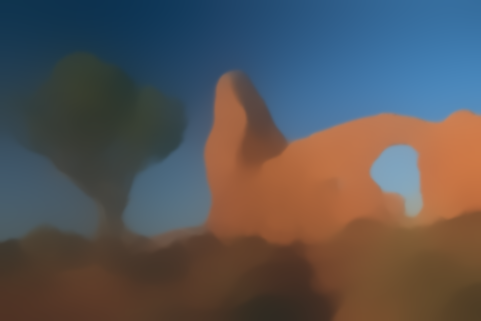}  \\
		original image \cite{Ber03} &
		$\alpha = 750$, $\beta = 50$ &
		$\alpha = 750$, $\beta = 500$ \\
		&
		\frame{
			\includegraphics[width=0.3\textwidth]
			{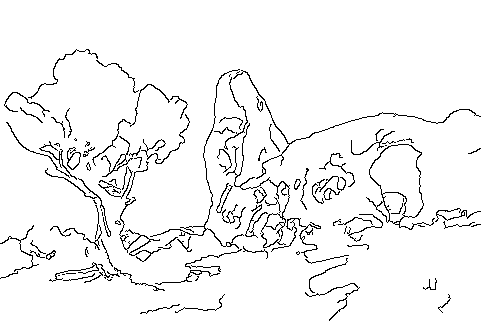}} &
		\frame{
			\includegraphics[width=0.3\textwidth]
			{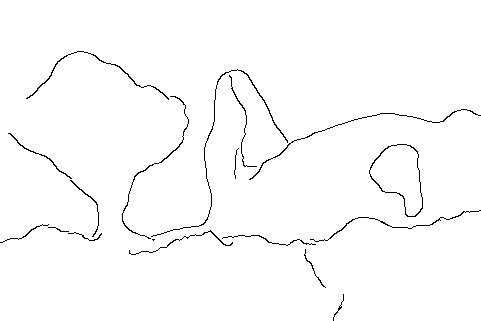}}
	\end{tabular}
	\caption{\textbf{First row:} original image and segmentation 
	results for our	first order	model.
		\textbf{Second row:} corresponding edge sets $\|\bm \nabla \bm u 
		- \bm v\|^2$.}
	\label{fig:figure4}
\end{figure}

In Fig. \ref{fig:figure4} we show our first order 
model for varying  values of $\beta$ together with the edge set that we 
get from the coupling term. For better visualisation we perform a 
hysteresis thresholding of the edge set \cite{Ca86}.
Our method is able to produce smooth segments. For an increasing
value of $\beta$, more and more segments are merged. For
instance we see that the smaller hole in the rock vanishes
for a higher value.
As already mentioned the resulting images are rather smooth in each 
segment due to the quadratic smoothness term, but we still have sharp edges 
in between. For an increasing $\beta$ value we obtain a sparser edge set.

\subsection{Application to Edge Detection}
\begin{figure}[tb]
	\centering
	\begin{tabular}{cc}
		\includegraphics[width=0.45\textwidth]
		{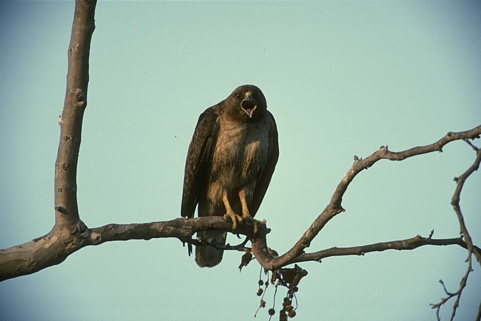} & 
		\includegraphics[width=0.45\textwidth]
		{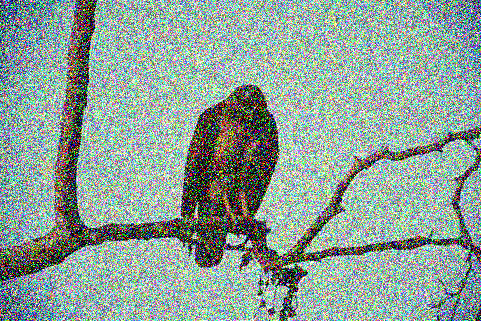} \\[2ex]
		\frame{
			\includegraphics[width=0.45\textwidth]
			{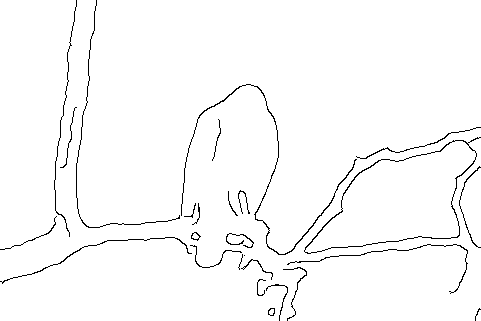}}&
		\frame{
			\includegraphics[width=0.45\textwidth]
			{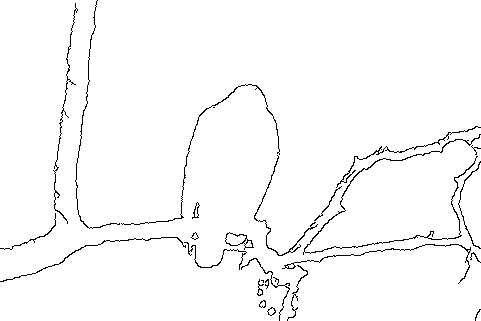}}
	\end{tabular}
	\caption{\textbf{First row:} Original image \cite{Ber03};
	noisy version ($\sigma=100$) 
  \textbf{second row:} edge set of the Canny edge detector;
	edge set of our first order model.}
	\label{fig:figure5}
\end{figure}

The previous subsection shows that an edge detector can be obtained 
directly from the coupling term. In Fig. \ref{fig:figure5} we
illustrate its high robustness under noise. We compare it to a Canny
edge detector \cite{Ca86}, which is a classical edge detection
method. 
To compensate for the noise, the Canny edge detector presmoothes the 
image by a convolution with a Gaussian kernel. 
While most large scale structures are detected, edges of smaller 
details are dislocated or merged together. This can be seen e.g.
at the branch below the bird. Our method 
works edge-preserving and is able to capture these small details.

\section{Conclusions and Outlook}
We have shown that coupling terms in variational models are 
useful for more than just easing numerics.
Already first order models that contain a coupling term in the 
sparsity promoting $L^1$ norm are sufficient for edge-preserving 
denoising. For the remaining data and smoothness terms we can 
use simple quadratic $L^2$ norms. This does not only simplify the 
numerical minimisation scheme, it also challenges the common belief 
that edge-preserving denoising requires a nonquadratic smoothness
term. 
The first order coupling model also shows a theoretical connection to 
the classical segmentation model of Mumford and Shah. 
As it works on a strictly convex energy, we obtain an approximation
of the Mumford-Shah functional that has a unique solution and is
guaranteed to converge.
Furthermore, the introduced coupling term itself carries valuable 
information: It has the properties of a global edge detector that is 
robust against noise and comes at literally no cost as a byproduct of 
the minimisation.

In our ongoing research we investigate how the coupling formulation
can be extended to applications in deblurring or inpainting or in
combination with anisotropic regularisation.
Moreover, we intend to exploit the simplicity of the quadratic terms 
by using parallel implementations to gain more 
efficient solvers.

\section*{Acknowledgements}
This project has received funding from the European Research Council 
(ERC) under the European Union’s Horizon 2020 research and innovation
programme (grant agreement no. 741215, ERC Advanced Grant INCOVID).


\bibliographystyle{abbrv}
\bibliography{myrefs}

\end{document}